%% file: main.tex
\documentclass[letterpaper, 10 pt, conference]{ieeeconf}  

\IEEEoverridecommandlockouts                              

\usepackage{cite}
\usepackage{amsmath}
\usepackage{amssymb}
\usepackage{soul}
\usepackage{booktabs}
\usepackage{makecell}
\usepackage{siunitx}
\usepackage{tabularray}
\usepackage{adjustbox}
\usepackage{pgfplots}
\pgfplotsset{compat=1.17}
\usepackage{tikz}
\usepackage{graphicx}
\usepackage{subcaption}
\usepackage{hyperref}
\usepackage{rotating}
\usepackage{multirow}
\usepackage[normalem]{ulem}
\usepackage{comment}

\usepackage{adjustbox} 

\newcommand{\spr}[1]{\textcolor{red}{#1}}

\newcommand{\raj}[1]{\textcolor{blue}{#1}}

\UseTblrLibrary{booktabs,siunitx}

\title{{\LARGE \bf PathFinder: Attention-Driven Dynamic Non-Line-of-Sight Tracking with a Mobile Robot}}

\author{Shenbagaraj Kannapiran$^{\dagger 1}$, Sreenithy Chandran$^{\dagger 2}$, Suren Jayasuriya$^{2}$, and Spring Berman$^{1}$
\thanks{$\dagger$ Equal contribution}
\thanks{This work was funded 
by 
NSF Award IIS-1909192. The authors acknowledge Research Computing at
Arizona State University for providing GPU resources for this research.}
\thanks{$^{1}$Shenbagaraj Kannapiran and Spring Berman are with the School for Engineering of Matter, Transport and Energy, Arizona State University, Tempe, AZ 85287, USA
        {\tt\small shenbagaraj@asu.edu, spring.berman@asu.edu}}
\thanks{$^{2}$Sreenithy Chandran and Suren Jayasuriya are with the School of Electrical, Computer and Energy Engineering, Arizona State University, Tempe, AZ 85281, USA~
        {\tt\small schand56@asu.edu, sjayasur@asu.edu}}%
}
\begin{document}
\maketitle
\thispagestyle{empty}
\pagestyle{empty}
\input{sec/0_abstract}
\input{sec/1_intro}

\input{sec/3_approach}

\input{sec/4_dataset}
\input{sec/5_experiments}

\input{sec/6_conclusion}
{
    \small
    \bibliographystyle{IEEEtran}
    \bibliography{main}
}

\end{document}

%% file: sec/0_abstract.tex
\begin{abstract}
The study of non-line-of-sight (NLOS) imaging is growing 
due to its many potential 
applications, including rescue operations and pedestrian detection by self-driving cars. However, implementing 
NLOS imaging on 
a moving camera 
remains an open area of research. Existing NLOS imaging methods rely on time-resolved detectors and laser configurations that require precise optical alignment, making it difficult to deploy them in dynamic environments. This work proposes a data-driven approach to NLOS imaging, PathFinder, that can be used with a standard RGB camera mounted on a small, power-constrained mobile robot, such as an aerial drone. Our experimental pipeline is designed to accurately estimate  the 2D 
trajectory of a 
person who moves in a Manhattan-world environment while remaining hidden from the camera's field-of-view. We introduce a novel approach to process a sequence of dynamic successive frames in a line-of-sight (LOS) video using an attention-based neural network that performs inference in real-time. The method also includes a preprocessing selection metric that analyzes images from a moving camera which contain multiple vertical planar surfaces, such as walls and building facades, and extracts planes that return maximum NLOS information. We validate the approach on in-the-wild scenes using a drone for video capture, thus demonstrating low-cost NLOS imaging in 
 dynamic capture environments.
\end{abstract}

%% file: sec/1_intro.tex
\section{Introduction}
\label{sec:intro}

Non-line-of-sight (NLOS) imaging is a technique that reconstructs an object (the ``NLOS object'') that is not in the direct line-of-sight (LOS) of a camera, using light scattered from one or more surfaces near the occluded object. This light undergoes multiple reflections and scatterings before it reaches a detector or camera, resulting in a low signal-to-noise ratio. 
To overcome this, a combination of optical setups, powerful detectors, imaging algorithms, or deep learning techniques can be used to estimate the underlying NLOS signal~\cite{geng2021recent,maeda2019recent}.
This method has 
a variety of potential 
applications, such as in medical imaging, autonomous driving (e.g., detecting pedestrians and other vehicles around corners), localization of disaster victims, and search-and-rescue operations in hazardous environments~\cite{borges2012pedestrian, geng2021recent, maeda2019recent}. 

\begin{figure}[t!]
  \centering
\includegraphics[width=\linewidth]{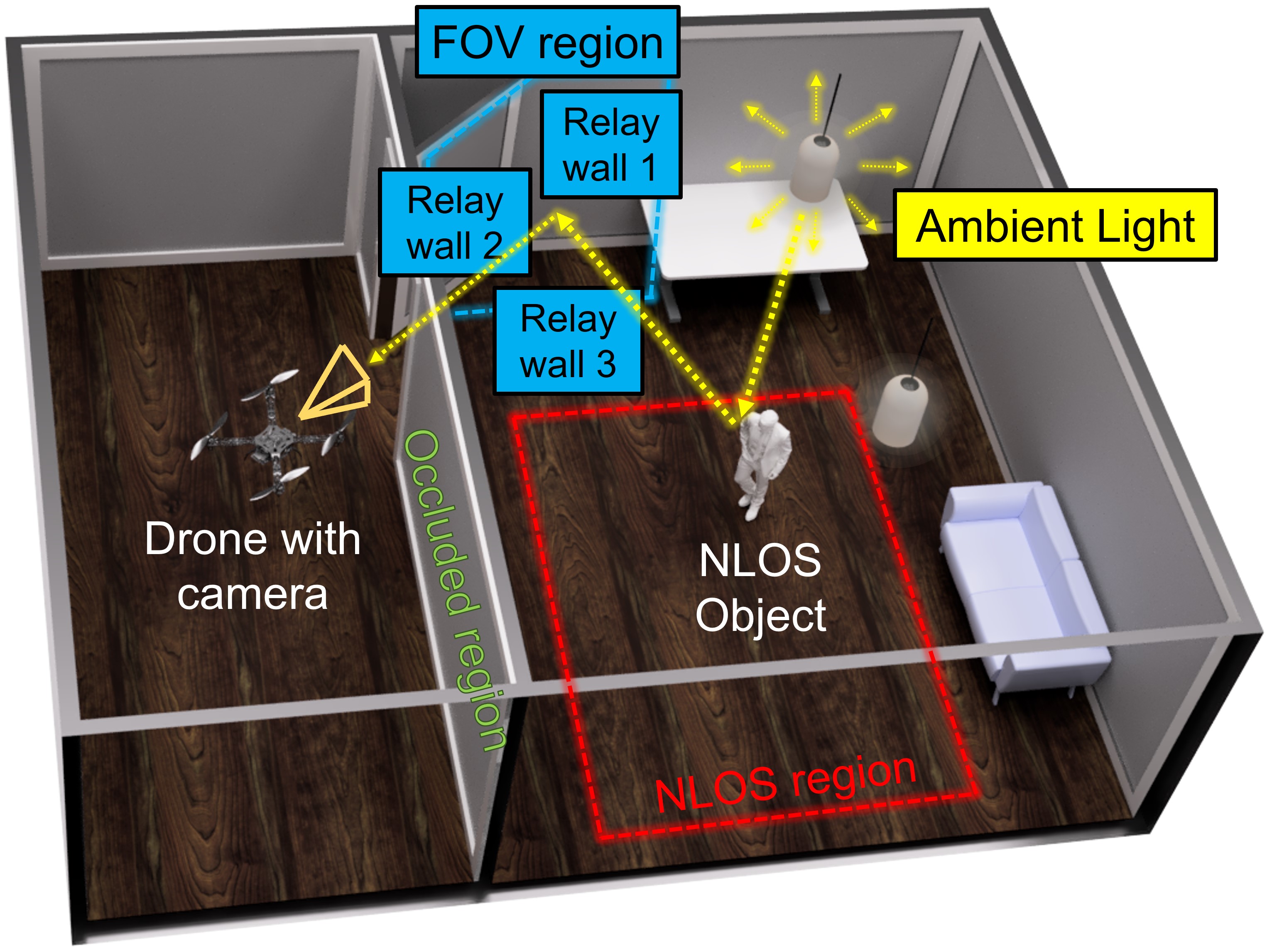}
  \caption{
  NLOS imaging task addressed by our method, which 
  estimates a 
  person's 2D trajectory 
  by leveraging the light scatter information in a drone's capture of several relay walls.} 
  \label{fig:teaser}
  \vspace{-4mm}
\end{figure}

Most NLOS imaging demonstrations tend to be restricted to laboratory-scale setups, with minimal or no movement of the detector/acquisition system. However, for NLOS imaging to be deployed robustly in practice, it needs to accommodate dynamic movements of the detector,  e.g., when it 
is mounted on a mobile robot, and work in large-scale environments. Thus far, few works have addressed this problem,
and existing approaches often utilize a portable radar sensor as the moving detector, 
e.g.,~\cite{rabaste2017around,scheiner2020seeing}. In an effort to fill this gap, we develop a low-cost, practical solution to 
NLOS imaging, {\it PathFinder}, that can be deployed in dynamic capture environments using conventional consumer cameras, without the need for specialized detectors.

NLOS imaging methods can be categorized as {\it active}, which make use of active illumination, or {\it passive}, which do not.
Active imaging techniques usually direct a high-temporal-resolution light source (e.g., pulsed laser) into the NLOS region and use a time-resolved detector, such as a streak camera~\cite{velten2012recovering} or Single Photon Avalanche Diodes (SPADs)~\cite{buttafava2015non, wu2021non}, to calculate the time of arrival of the reflected light pulse~\cite{kirmani2011looking, tsai2017geometry, xin2019theory}. Since these methods acquire very precise time data, they are suitable for high-resolution 3D object reconstruction. However, they can only be implemented using elaborate optical setups and require long acquisition times. In state-of-the-art Time-of-Flight systems, the scanning frequency can take up to several minutes, which is insufficient for real-time NLOS applications~\cite{o2018confocal, lindell2019wave}. In contrast to time-of-flight, researchers have also explored NLOS imaging using conventional cameras and lasers~\cite{klein2016tracking, chen2019steady} and/or spotlight illumination~\cite{chandran2019adaptive,chandran2024learning}. 
However, this method still requires the use of a controlled illumination source, adding size, weight, and power when deployed on a robotic platform.

We adopt a passive NLOS imaging approach, which is more suitable for our goal. Passive NLOS imaging methods, e.g.,~\cite{baradad2018inferring, bouman2017turning, wang2023propagate, sharma2021you, krska2022double, cao2022computational, yedidia2019using, geng2021passive}, capture the visible light reflected from the hidden object to perform the imaging task.
Due to the ill-posed nature of the problem, additional constraints and priors are often applied, including partial occlusion~\cite{seidel2020two,torralba2014accidental}, polarization~\cite{yedidia2019using}, and coherence~\cite{beckus2019multi}.  Recently, the use of data-driven scene priors for passive NLOS imaging has shown great promise~\cite{tancik2018flash,aittala2019computational}. Tancik et al.~\cite{tancik2018flash} 
used a convolutional neural network (CNN) to perform activity recognition and tracking of humans and a variational autoencoder to perform reconstructions.  
Sharma et al.~\cite{sharma2021you} presented 
a deep learning technique that can detect the number of individuals and the activity performed by observing the LOS wall~\cite{sharma2021you}.  
Wang et al.~\cite{wang2023propagate} introduced PAC-Net, which utilizes both static and dynamic information about an NLOS scene. The method alternates between processing difference images and raw images to perform tracking. 
However, all of these methods are restricted to scenarios with a static camera.

Passive NLOS imaging methods are usually used for low-quality 2D reconstructions and localization tasks and often suffer from a low signal-to-noise ratio (SNR), a challenge previously addressed by subtracting the temporal mean of the video from each frame~\cite{sharma2021you, he2022non, bouman2017turning}. However, this background subtraction technique is not feasible in dynamic capture environments. To overcome this limitation, we propose a data preprocessing pipeline that enhances the SNR and facilitates scene understanding.
Moreover, unlike existing passive methods that estimate an object's position based on a single stationary planar surface, our approach accommodates scenarios where the camera, steered by a robot platform, captures varying sections of multiple planar surfaces. Recognizing that all visible surfaces could contain valuable NLOS scatter information, we develop 
a transformer-based network that leverages captures from all of these surfaces to estimate the position of a hidden NLOS object. We train our pipeline with a mixture of both synthetic and real data.


Our primary contributions are as follows:
\begin{itemize}[leftmargin=*]
    \item We introduce a novel approach to NLOS imaging with a moving camera, employing a vision transformer-based architecture that uses example packing to simultaneously process 
    multiple flat relay walls with different aspect ratios, thereby enhancing NLOS tracking performance.
    \item We demonstrate state-of-the-art results on real data to validate our method, using a quadcopter for video capture.
    \item We collect 
    the first dataset of  dynamic camera footage synchronized with high-resolution real NLOS object trajectories and camera poses, which we plan to release~\cite{dataset_link}.
\end{itemize}

%% file: sec/3_approach.tex
\section{Problem Statement}
Figure~\ref{fig:teaser} illustrates 
our imaging setup. 
A small mobile robot 
(here, a quadcopter) equipped with a forward-facing RGB camera 
moves in an occluded region while capturing images of multiple relay walls (here, viewed through an open door) that are within its field-of-view (FOV) region. 
The goal of our method is to estimate the time-varying 2D position of a person (NLOS object) outside the camera's FOV  as they walk around an area that is not visible to the robot (NLOS region). The yellow lines 
illustrate how light reflected from the relay walls contains scattered information from the NLOS object, which is captured by the camera on the robot.  

The raw image of a visible planar surface captured by the camera is denoted by $\mathbf{I} \in \mathbb{R}^2$,  which can be described as the output of a reflection function $\mathcal{F}$, 
$\mathbf{I} = \mathcal{F}(\mathbf{X}, \mathbf{V}, \mathbf{N}, \omega)$,
where $\mathbf{X} \in \mathbb{R}^2$ is the person's ground-truth 
position in a plane parallel to the floor; $\mathbf{V} \in \mathbb{R}^2$ is their ground-truth velocity in this plane;
$\mathbf{N} \in \mathbb{R}^3$ is the unit vector normal to the surface;
and $\omega$ is 
a set of environmental and material 
parameters that affect the appearance of the image, 
including ambient noise and surface reflectivity. The direction of $\mathbf{N}$ with respect to the NLOS object 
determines 
the amount of NLOS scatter information that reaches the surface. The function $\mathcal{F}$ models the light transport of the setup; that is, it contains information about how light interacts with the scene before reaching the camera. Our approach learns the inverse function $\mathcal{F}^{-1}$ in order to compute estimates of 
$\mathbf{X}(t)$ and $\mathbf{V}(t)$ for $t \in [0, T]$, given some final time $T$. We denote these estimates by $\mathbf{X}'(t)$ and $\mathbf{V}'(t)$, respectively. 
In this study, we assume that there is only a single occluded person and that the scene has a Manhattan-world configuration.


\section{NLOS Tracking Pipeline}
\begin{figure*}[t!]
	\centering
	\includegraphics[width=\linewidth]{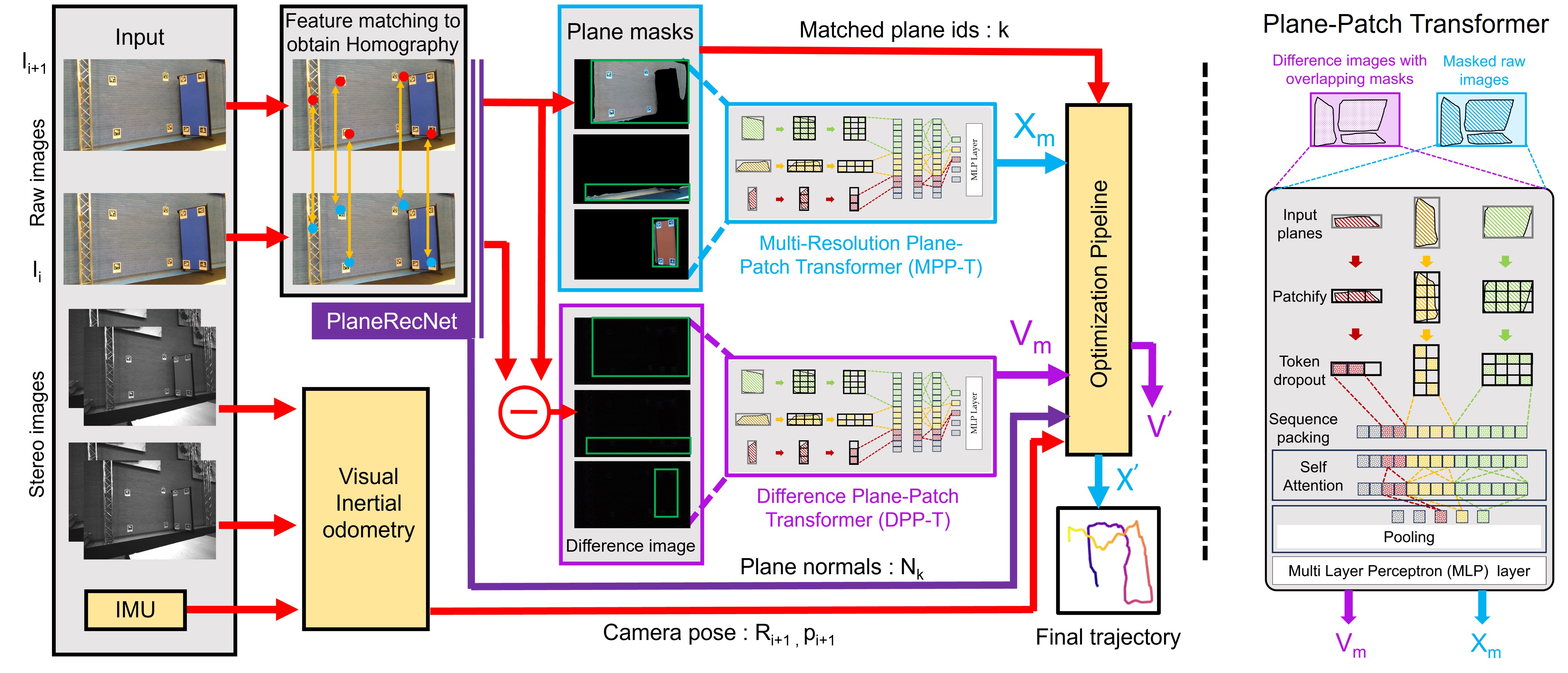}
	\caption{Inference pipeline for NLOS object tracking. 
 Raw images, stereo image pairs, and IMU data are input to VIO to estimate the camera pose. PlaneRecNet generates plane masks from consecutive images. 
 Homography from feature matching is applied to plane masks, creating difference images and plane IDs $k$. The raw image at time step $i+1$ and the difference image between time steps $i$ and $i+1$ are input to MPP-T and DPP-T networks, which compute 
 the estimates 
 $\mathbf{X}_m$ and $\mathbf{V}_m$ for each example plane $m$ (see Section \ref{nlospatchnet}). 
These estimates, 
 along with the camera pose, plane IDs, 
 and unit vector normal to each plane, 
 are input to 
 an optimization layer to compute 
 the NLOS object's 
 trajectory. The figure on the right shows details of the Plane-Patch Transformer architecture. 
 }
	\label{fig:fullpipeline}
\end{figure*}

In this section, we delineate the various stages of our proposed pipeline, depicted in Fig. \ref{fig:fullpipeline}. 
Given that the camera is moving, 
the pipeline commences with a plane extraction process, which is described 
in Section \ref{sec:PEP}. This process operates on the raw data stream alongside additional inputs from the capture system, yielding masked, separated planes that are instrumental in NLOS object tracking. Subsequently, the output of this pipeline feeds into the NLOS transformer network, which estimates $\mathbf{X}$ and $\mathbf{V}$ in different planes.
A description 
of the network architecture is provided in Section \ref{nlospatchnet}. We also refine the estimates of $\mathbf{X}$ and $\mathbf{V}$  using data from multiple planes, as detailed in Section \ref{sec:optimisation}. 

\subsection{Plane Extraction Pipeline}\label{sec:PEP}
We first use visual inertial odometry (VIO) to obtain pose estimates of the moving camera and feature matching to identify anchor points in the image. The pipeline utilizes raw RGB camera images, stereo image pair data, and inertial measurement unit (IMU) data as input. The ground-truth trajectory of the person is obtained from
motion capture data. We use the stereo pair images and synchronized IMU data to perform VIO  using the multi-state constant Kalman filter (MSCKF)~\cite{sun2018robust}, which provides  reliable camera pose estimates. When the camera is mounted on a drone, 
the VIO output is used to localize the drone while in flight. 
 
The VIO method is used to obtain the camera's positions $\mathbf{p}_i$ and $\mathbf{p}_{i+1}$ and 
orientations $\mathbf{R}_i$ and 
$\mathbf{R}_{i+1}$ in the global coordinate system at time steps $i$ and $i+1$, 
respectively. 
Consider the raw color images 
$\mathbf{I}_i$ and $\mathbf{I}_{i+1}$ that are captured at these times.  
The main function of the pipeline is to monitor all planar surfaces that can serve as intermediary relay walls to carry out effective NLOS object tracking by extracting image patches from these surfaces to pass along to the downstream NLOS-Patch network. To accomplish this, we use the current state-of-the-art learning-based plane detection algorithm PlaneRecNet~\cite{xie2021planerecnet}, which gives us the necessary plane detection masks. 
The network does not track planes or assign plane IDs. Instead, we use SIFT feature matching for plane tracking between consecutive images and their difference images, estimating the homography for stitching. Plane IDs are determined based on mask overlap in stitched images; new IDs are assigned to non-overlapping planes, and  IDs of non-visible planes are discarded.  The input to the transformer-based NLOS-Patch network includes the plane masks, plane IDs, raw color images, and difference images. 

\subsection{NLOS-Patch Network}\label{nlospatchnet}
The NLOS-Patch network is designed to process the raw images and compute a 2D position estimate $\mathbf{X}_m$ and 2D velocity estimate $\mathbf{V}_m$ of the NLOS object in 
each 
plane $m$ in a sequence of $M$ example planes (see Section \ref{sec:masked-self-attention}), represented in the camera coordinate system. These estimates are for time step $i+1$, with $\mathbf{X}_m$ updated based on the assumption that $\mathbf{V}_m$ at time step $i$ is constant over the small time interval between time steps. As shown in Fig.~\ref{fig:fullpipeline}, this parallel transformer network consists of a Multi-resolution Plane-Patch Transformer (MPP-T), which computes $\mathbf{X}_m$, 
and a Difference Plane-Patch Transformer (DPP-T), which computes $\mathbf{V}_m$, $m=1,...,M$. 
The MPP-T and DPP-T are 
vision transformers (ViTs)~\cite{dosovitskiy2020image}, which have been shown to have advantages over convolutional networks for a variety of computer vision tasks~\cite{carion2020end, kirillov2023segment}. 
In a ViT, an image is divided into patches, 
and each patch is linearly transformed into 
a token. These tokens are then passed into attention modules for learning.  Generally, the input is reshaped into a square matrix and then divided into a fixed number of grids. However, this input reshaping and subdivision is highly inefficient for our purpose, since the visual feed of a mobile robot will vary with each successive observation as it 
navigates the environment, 
given its restricted FOV. 
Thus, our NLOS-Patch network should determine $\mathbf{X}_m$ and $\mathbf{V}_m$ 
using incomplete images of planes and patches of various sizes and locations. 

Additionally, since our passive NLOS tracking approach already suffers from low SNR due to the presence of ambient lighting, our goal is to leverage all available spatial intensity information  captured from the LOS surfaces in our learning problem. Toward this end, we design our transformers based on the NaViT network proposed by Dehgani et al.~\cite{dehghani2023patch}, which packs multiple patches from different images with varying resolutions into a single sequence. They 
demonstrate that example packing, wherein multiple examples are packed into a single sequence, results 
in improved performance and faster training. This is a popular technique in natural language processing.
The main components of the 
network 
are described below.

\subsubsection{Patchify}
Given a raw image $\mathbf{I}_i$ at time step $i$, a detected plane in $\mathbf{I}_i$ with ID $k$, and a corresponding plane mask image denoted by $\mathbf{M}_{i,k}$, a masked plane $\mathbf{Im}_{i,k}$ is obtained as the product of $\mathbf{I}_i$ and $\mathbf{M}_{i,k}$. The
 masked planes $\mathbf{Im}_{i,k}$ and $\mathbf{Im}_{i+1,k}$ at time steps $i$ and $i+1$ and the difference image $\Delta \mathbf{Im}_{i+1,k}$ between them are passed into the NLOS-Patch network. The difference image $\Delta \mathbf{Im}_{i+1,k}$ can be used to estimate the NLOS object's velocity $\mathbf{V}$ between time steps $i$ and $i+1$, as demonstrated by Wang et al.~\cite{wang2023propagate}.
The masked planes at each time step 
are then packed into a single batch, and 
each image in this batch is split into patches. Then, we apply token dropout to the patches and obtain the resulting sequences of masked planes. This procedure is also applied to the difference images and plane IDs. 


\subsubsection{Factorized Positional Embedding} 
To process images of arbitrary resolutions, we use factorized absolute position embeddings $\phi_w$ and $\phi_h$ for the width and height of the patches, respectively, as proposed in~\cite{dehghani2023patch}. These embeddings are each summed with the learned patch embedding, $\phi_p$.

\subsubsection{Masked Self-Attention} \label{sec:masked-self-attention}
We employ self-attention masks to learn attention relationships between planes with identical IDs.
Masked pooling at the end of the attention layers ensures that token representations are pooled within each example. The output of this pooling is a single vector 
for each of the $M$ example planes in the sequence, which is finally passed into a simple Multi-Layer Perceptron (MLP) head consisting of two fully-connected layers. The MLP network outputs the estimates $\mathbf{X}_m$ and $\mathbf{V}_m$, $m=1,...,M$. 

\subsubsection{Loss Functions}
For end-to-end training of our NLOS-Patch network, our objective is to minimize the error in estimating the position and velocity of the NLOS object. To achieve this, we define 
a loss function $\mathcal{L}$ that takes into account the estimates from all $M$ example planes, so that the network learns from the diverse information provided by multiple reflective planes: 
\begin{equation*}
  \mathcal{L} = \sum_{m=1}^M \left( \text{MSE}(\mathbf{X},\mathbf{X}_m) + \alpha \hspace{1mm} \text{MSE}(\mathbf{V}, \mathbf{V}_m) \right),
\end{equation*}
 where MSE is the mean squared error and $\alpha$ is a constant weighting parameter that balances the relative importance of the position and velocity losses. 

\subsection{Optimization Pipeline} \label{sec:optimisation}


The estimates $\mathbf{X}_m$ and $\mathbf{V}_m$ in each example plane $m$ at time step $i+1$ are 
passed from the NLOS-Patch network into the optimization pipeline to produce the final estimates $\mathbf{X}'$ and $\mathbf{V}'$. For each example plane $m$, the estimates $\mathbf{X}_m$ and $\mathbf{V}_m$ are transformed from the camera coordinate system to the global coordinate system using the transformation matrix $\mathbf{T}_m \in \mathbb{R}^{3 \times 4}$ corresponding to the plane.  We denote the estimates in global coordinates as $\mathbf{X}_m^g$ and $\mathbf{V}_m^g$.
Since we model the person as moving 
in the 
$(x,y)$ plane of the global coordinate system (see Fig. \ref{fig:dataset1}(b)), we set both the $z$ coordinate of $\mathbf{X}_m^g$ and the $\dot{z}$ coordinate of $\mathbf{V}_m^g$ to 0.
Let $m_1$, $m_2$, and $m_3$ be the indices of the largest, second-largest, and third-largest example planes, respectively. We denote the reflections of the position estimates $\mathbf{X}_{m_2}^g$ and $\mathbf{X}_{m_3}^g$ 
 across planes $m_2$ and $m_3$, respectively, by $\mathbf{X}_{m_2}^{g,r}$ and $\mathbf{X}_{m_3}^{g,r}$. These reflected position estimates (in global coordinates) are computed as the output of a reflection function $\mathcal{F}_{\text{g}} (\mathbf{X}_m^g, \mathbf{V}_m^g, \mathbf{N}_m, \mathbf{T}_m)$, $m \in \{m_2, m_3\},$ where \( \mathbf{N}_m \) is the unit normal vector of example plane $m$. The reflection function is the composition of a series of operations that model how light interacts with the NLOS object 
and a reflective plane before reaching the camera. Our optimization problem computes the position estimate $\mathbf{X}'$ and velocity estimate $\mathbf{V}'$ that minimize the following cost function:
\begin{equation} 
    J(\mathbf{X}', \mathbf{V}') = \hspace{-5mm} \sum_{m \in \{m_2,m_3\}} \left\|  \mathbf{X}_{m_1}^g - \mathcal{F}_{\text{g}}(\mathbf{X}^g_m, \mathbf{V}^g_m,  \mathbf{N}_m, \mathbf{T}_m) \right\|_2^2. \nonumber
\end{equation}

%% file: sec/4_dataset.tex
\section{Datasets for NLOS-Patch Network Training}
\label{sec:dataset}




\subsection{Synthetic Dataset}
\begin{figure}[t!]
	\centering
	\includegraphics[width=\linewidth]{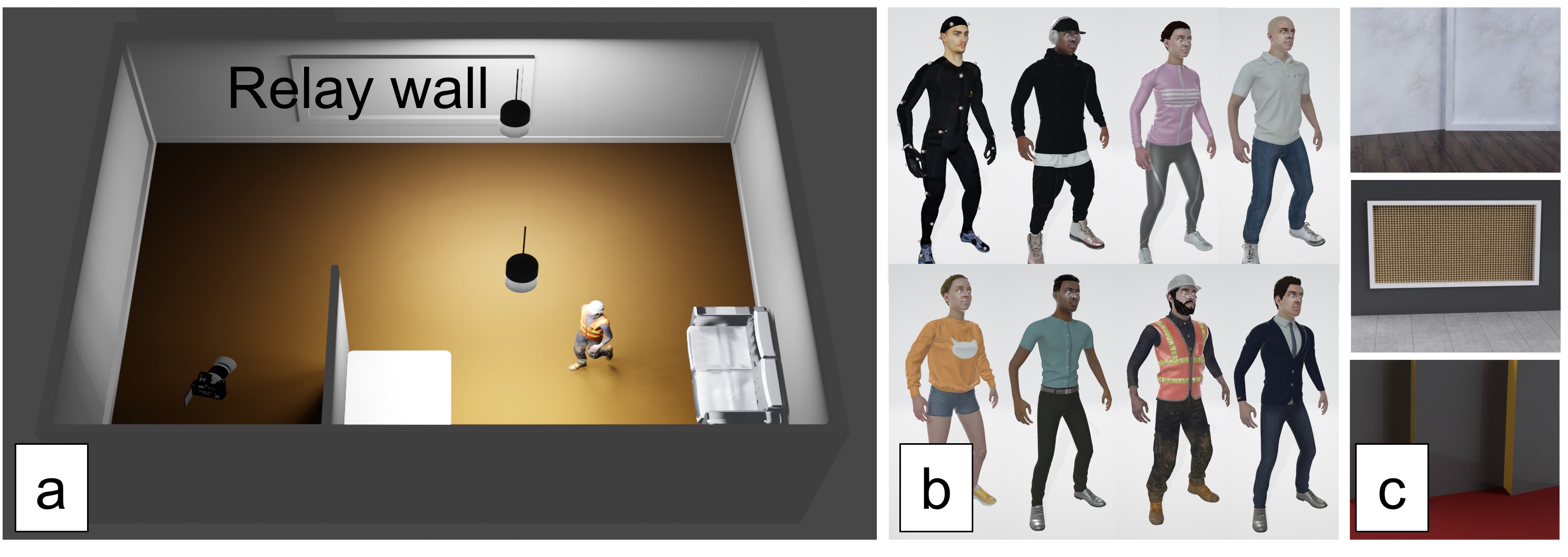}
	\caption{(a) Overhead view of a sample synthetic NLOS scene simulated using Blender, showing the 
 camera  ({\it lower left}), 
  human character (NLOS object), and 
 sources of ambient lighting in the room. (b) Samples of the eight characters from the Mixamo library that were used for synthetic data generation. (c) Samples of three sets of relay walls with different materials that were used for synthetic data generation.}
	\label{fig:sim}
\vspace{-5mm}
\end{figure} 

We generated synthetic data to train the NLOS-Patch network using Cycles,
Blender's physics-based ray tracing renderer. 
We simulated diverse indoor NLOS imaging scenarios in an effort to create to a rich training dataset that enhances the network's robustness
in complex real-world scenarios. In each simulation, the camera moved along a random trajectory in 3D space, and the NLOS object was an animated human character, created in Mixamo, who walked with different gaits and postures in a 
random path along the ground plane. 
Trajectories for the 6D camera pose and the 2D position of the human character were each defined by selecting 10 random points in a specific region and connecting them with a B\'ezier curve. To increase the variance in the dataset, we simulated 20 different configurations of objects in the scene, eight different human characters, and different numbers, orientations, and materials of  relay walls in the FOV region (see Fig. \ref{fig:sim}).
To enhance the realism of the synthetic data, we also simulated real-world noise at the pixel level 
in the synthetic data creation process. 

The Robot Operating System (ROS) was integrated with Blender scripts through the use of Blender addons, 
enabling easy collection of the trajectories of the human character and camera 
from their respective ROS topics. The rendering for frames of size 256 $\times$ 256 pixels 
took around 3 s per frame. 

\subsection{Real-World Dataset and Hardware Configuration}\label{realdata}

We also trained the NLOS-Patch network using real-world data, which we collected from 
trials with 5 human subjects across 10 large-scale indoor scenes. We plan to publicly release this dataset 
\cite{dataset_link} for researchers in the community.

\begin{figure}[t!]
  \centering
  \includegraphics[width=\linewidth]{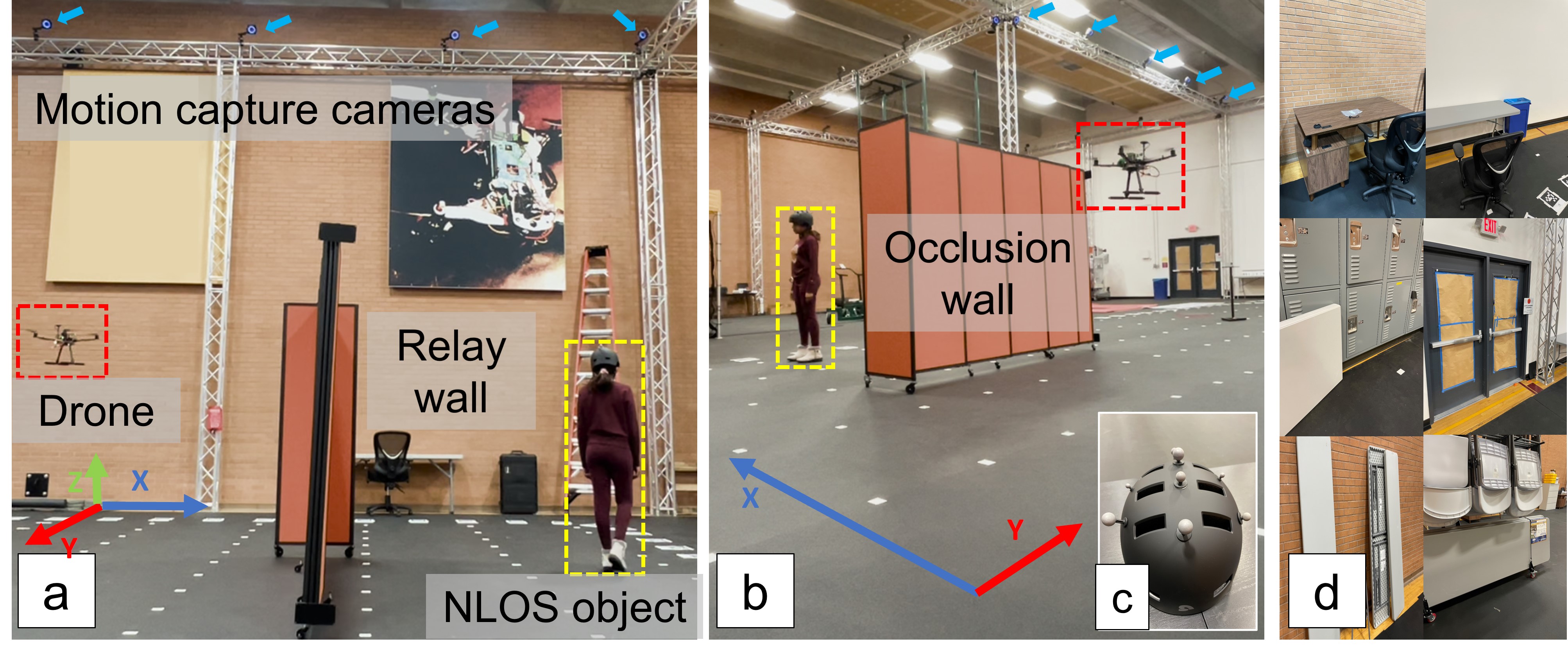}
  \caption{(a) Real-world data collection setup: A drone captures images of a relay wall while a person (NLOS object) is hidden from view. The person's ground-truth position is obtained using motion capture cameras. (b) Side view of the 
  setup. 
  (c)  Helmet mounted with IR markers for ground-truth data collection. 
  (d) Samples 
  of 
  FOV regions in the dataset, with different 
 surface textures and types of objects present.}
  \label{fig:dataset1}
\end{figure}

\begin{figure}[t!] 
  \centering
  \includegraphics[width=0.7\linewidth]{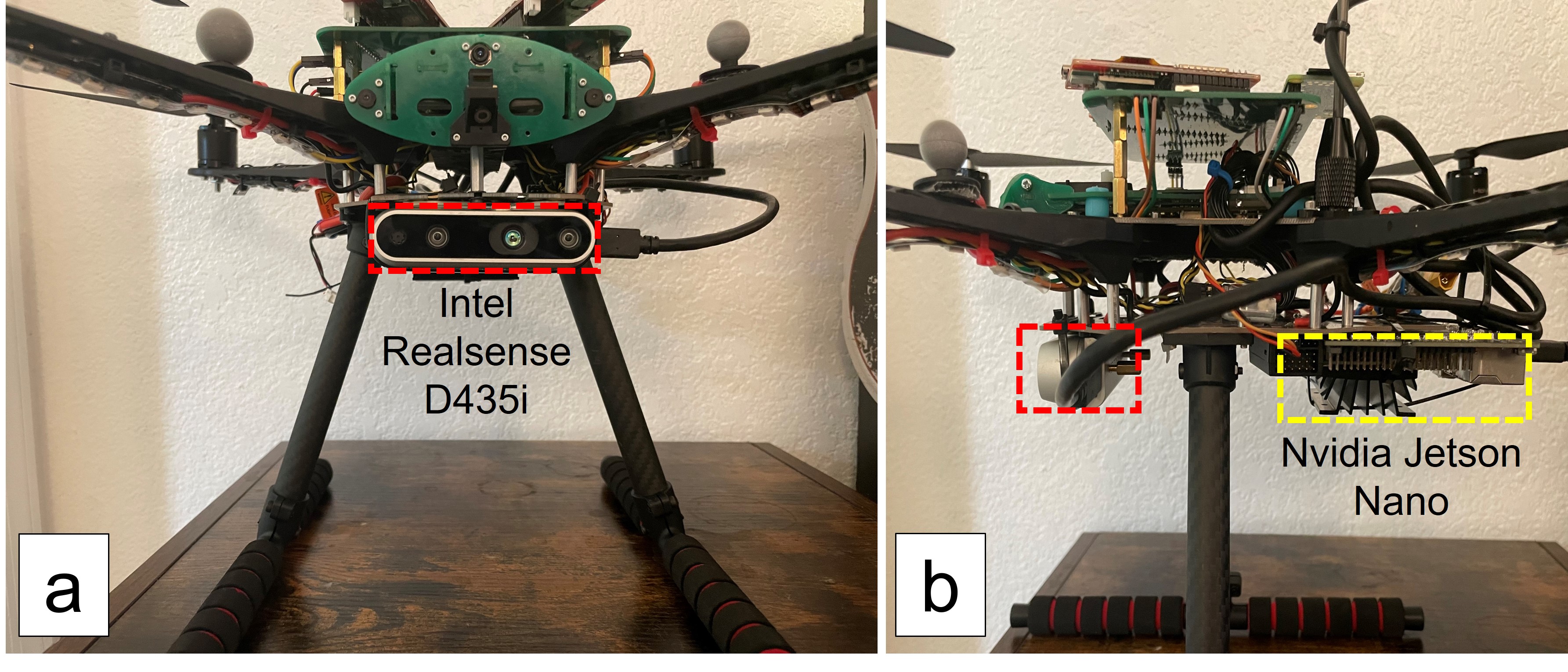}
  \caption{Customized drone equipped with Intel RealSense cameras ({\it highlighted in red}) for visual inertial odometry (VIO) during indoor flight. Raw data, including color camera images, stereo images, and IMU data, are extracted for VIO and camera pose estimation. The onboard Jetson Nano ({\it highlighted in yellow}) facilitates real-time processing.}
  \label{fig:dataset2}
\end{figure}

The data collection setup is shown in Figs.~\ref{fig:dataset1}(a)-(b), with images of sample FOV regions shown in Fig.~\ref{fig:dataset1}(d). We built a versatile aerial drone (Fig.~\ref{fig:dataset2}) 
using standard off-the-shelf components to serve as the mobile camera platform. The drone is equipped with 
an Intel RealSense depth camera D435i, which has a 2-MP RGB camera with a resolution of 1920 × 1080 pixels and a FOV of 69° × 42°. The camera operates with a rolling shutter mechanism and can capture images at a rate of 30 FPS. 
The drone is also equipped with
a stereo pair of cameras with a resolution of 1280 × 720 pixels and a combined FOV of 87° × 58°, which can capture images at 90 FPS. 
The onboard synchronized IMU 
is also integrated into our data collection methodology, providing visual inertial odometry.

The indoor testing space 
was equipped with 68 OptiTrack Prime 17W motion capture cameras with a 70°-degree horizontal FOV and a 1.7-MP (1664 $\times$ 1088 pixel resolution) image sensor, which captured position data at a rate of 120 FPS with $<0.5$ mm precision. 
We used this motion capture system to track infrared (IR) reflective markers that were attached to the drone and to a helmet worn by the participants (Fig. \ref{fig:dataset1}(c)). In this way, we obtained precise ground-truth data for the occluded person's
positions and the drone's poses. 



To collect the images captured by the cameras onboard the drone, the Intel RealSense camera was connected to an NVIDIA Jetson Nano computer on the drone, which recorded the raw data onto the NVMe SSD storage. The data were then synchronized with ROS 
and extracted after the trials as ROSbag files for further processing. The data collection was done offline because of significant latency in transmitting the raw camera and stereo pair image data in real-time over WiFi, 
due to the bandwidth requirements (250 Mb/s). 

%% file: sec/5_experiments.tex
\section{Experiments}
\label{sec:experiments}
In this section, we describe 
the NLOS-Patch network training process, the metrics used to evaluate the tracking performance of our method, and other ablations. We quantify and discuss our method's performance on our collected synthetic and real-world datasets. The supplemental video shows tracking results for our method in real-world experiments.

\subsection{Training Procedure} 

The NLOS-Patch network was trained with both synthetic and real data, and inference was 
done on both types of data. 
The entire dataset was split into non-overlapping training and validation datasets. 
Although the Intel RealSense camera streams at 30 FPS, 
the person (NLOS object) does not move to 30 different positions within a 1-second time interval, so we chose every 15$^{th}$ frame for our datasets.  
During the training step, we passed the three largest planes generated by the plane extraction pipeline (Section~\ref{sec:PEP}) into the network.
All values of the NLOS object's position were normalized by the size of the room. We employed a transformer network with a patch size of 64, an attention dimension size of 1024, a depth of 4, and a dimension head of 128. Additionally, we set the dropout rate for the tokenized input to 0.4 and the embedded dropout to 0.2.

{\it Inference Speed:} The PlaneRecNet algorithm and our VIO method both have an inference speed of 8-9 FPS. For an instance of processing three planes in a single sequence, the NLOS-Patch network has an inference time of 3000 FPS on an NVIDIA RTX A6000 graphics card. 
Hence, our NLOS tracking method is capable of real-time inference.

\subsection{Quantitative Tracking Results}

To assess the tracking performance of our method, we measured the Root Mean Square Error (RMSE) between the NLOS object's ground-truth position $\mathbf{X}(t)$ and its estimate $\mathbf{X}'(t)$, denoted by $\textsc{Rmse}_x$, and the RMSE between the NLOS object's ground-truth velocity $\mathbf{V}(t)$ and its estimate $\mathbf{V}'(t)$, denoted by $\textsc{Rmse}_v$. 
We compared the performance of our method on both synthetic and real-world data to that of the following other passive, deep learning-based NLOS imaging methods as baselines. Table~\ref{results} reports the resulting RMSE values (average $\pm$ standard deviation) over 50 trials of duration 128 s each. We note that our method is the first passive NLOS tracking method that uses a dynamic camera and multiple relay walls, whereas the selected baseline methods use a stationary camera and a single relay wall. \newline
\indent \textbf{PAC-Net (Wang et al.~\cite{wang2023propagate}):} 
Comparing our method to 
PAC-Net helps to assess 
the advantages of our transformer-based architecture and attention mechanism in handling dynamic camera scenarios.
Similar to our method, PAC-Net 
alternately processes raw images and difference images using two recurrent neural networks.
However, this method is trained on single static planar walls. We trained the PAC-Net network to run inference using the full captures of raw images in our input dataset. As shown in Table~\ref{results}, PAC-Net yielded an 
average  $\textsc{Rmse}_x \approx$ 55 mm and average $\textsc{Rmse}_v \approx$ 1.2 mm/s. 
\newline
\indent \textbf{He et al.~\cite{he2022non}:} 
This method uses a deep learning-based approach to image and track moving NLOS objects using RGB images captured under ambient illumination. It employs a CNN 
architecture followed by fully-connected layers. We re-implemented their proposed network architecture and trained it on our input dataset. This method produced an average $\textsc{Rmse}_x \approx$ 73 mm. \newline
\indent \textbf{Tancik et al.~\cite{tancik2018flash}:} 
This method uses a CNN regression network that is designed to learn from the scattered light information in the environment to achieve 
NLOS object tracking and activity recognition. 
This method produced an average $\textsc{Rmse}_x \approx$ 93 mm.

Compared to these baseline methods, our method without any modifications (the ``All patches'' rows in Table~\ref{results}) produced the lowest average $\textsc{Rmse}_x$ for both real-world data ($\sim$16 mm) and synthetic data ($\sim$18 mm). $\textsc{Rmse}_v$ values could not be computed for the He et al. and Tancik et al. methods because their network architectures only process raw images, not difference images. 

We also tested the performance of our method with the following modifications:
\newline
\indent \textbf{w/o velocity:} In this version of our method, we trained 
the transformer network without including the velocity loss $MSE(\mathbf{V},\mathbf{V}')$ in the loss function. 
Thus, there is a single transformer architecture that processes the masked  planes, and it is trained by minimizing  the position loss $MSE(\mathbf{X},\mathbf{X}')$. The average $\textsc{Rmse}_x$ for this method was $\sim$22 mm for real-world data and $\sim$27 mm for synthetic data.\newline 
\indent \textbf{w/o optimization:} This version of our method does not perform the optimization step that is critical to ensure accurate position estimates. Instead, the estimated position $\mathbf{X}'$ is computed as the average of the position estimates for the three largest planes output by the plane extraction pipeline 
across the patches in each example sequence in the NLOS-Patch network.
The average $\textsc{Rmse}_x$ for this version is significantly higher than that for the unmodified version, for both the synthetic and real-world data. \newline
\indent \textbf{One Patch:} In this version of our method, only the largest plane output by the plane extraction pipeline is passed into the transformer network. 
This version yields slightly worse performance in terms of average $\textsc{Rmse}_x$ than the unmodified version for both the synthetic and real-world data, demonstrating
the advantage of processing multiple planes simultaneously in our method.


\begin{table}[t!]
\centering
\footnotesize 
\begin{tabular}{cccc}
\toprule
\multicolumn{2}{l}{\textbf{}} & \textbf{\textsc{Rmse}$_x$} (mm) & \textbf{\textsc{Rmse}$_v$} (mm/s)\\
\toprule

\multirow{3}{*}{Baselines} & PAC-Net & 54.73 $\pm{24.78}$ & 1.22 $\pm{0.42}$ \\
 & He et al. & 72.65 $\pm{53.32}$ & - \\
 & Tancik et al. & 92.97 $\pm{64.87}$ & - \\
 \cmidrule(lr){1-4}
 
\multirow{4}{*}{\shortstack{Ours -\\PathFinder\\ (Real Data)}} & w/o velocity & 21.53 $\pm{2.64}$ & - \\
 & w/o optimization & 29.14 $\pm{2.79}$ & 1.34 $\pm{0.53}$\\
 & One patch & 16.72 $\pm{3.09}$ & \textbf{1.13 $\pm{0.46}$}\\
 & All patches & \textbf{15.94 $\pm{2.38}$} & 1.38 $\pm{0.34}$\\

  \cmidrule(lr){1-4}
\multirow{4}{*}{\shortstack{Ours -\\PathFinder\\ (Synthetic \\ Data)}} & w/o velocity & 26.75 $\pm{2.93}$ & - \\
 & w/o optimization & 27.93 $\pm{1.92}$ & 1.39 $\pm{0.33}$\\
 & One patch & 19.98 $\pm{2.24}$ & 1.43 $\pm{0.38}$\\
 & All patches & \textbf{18.05 $\pm{2.03}$} & \textbf{1.26 $\pm{0.39}$}\\
 
\bottomrule
\end{tabular}
\caption{
Tracking performance of our method 
(PathFinder), with and without modifications, and baseline methods. 
}
\label{results}
\vspace{-5mm}
\end{table}

\subsection{Qualitative Visualization of Tracking Performance}

Figure~\ref{fig:trajectory} plots several trajectories $\mathbf{X}'(t)$ estimated by our method against the corresponding ground-truth trajectories $\mathbf{X}(t)$, with the $\textsc{Rmse}_x$ indicated along the estimated trajectories. 
These plots 
illustrate how closely the estimated path from our method follows the actual path of the NLOS object. 

\begin{figure}[t!]
\vspace{-6mm}
  \centering
  \begin{subfigure}{\linewidth}
    \centering
    \includegraphics[width=0.99\linewidth]{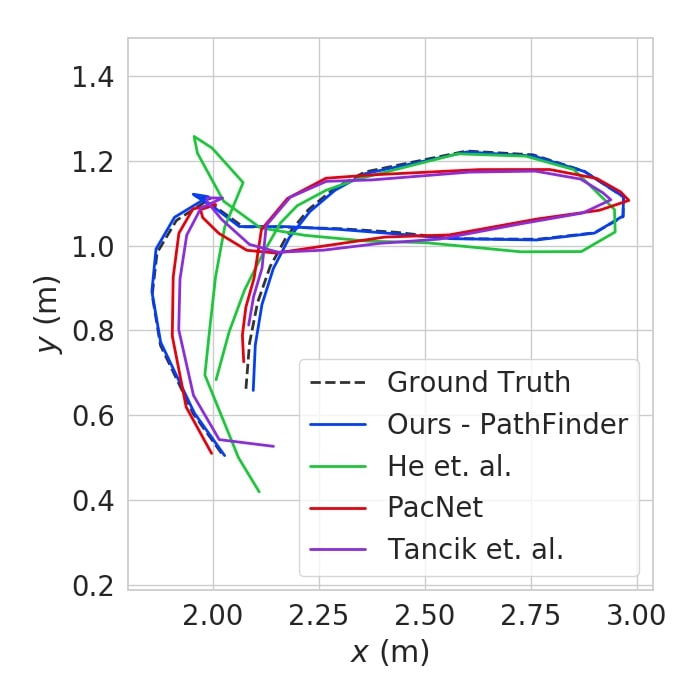}
    \label{fig:sub1}
  \end{subfigure}
  
  \vspace{-3mm} 
  
  \begin{subfigure}{0.39\linewidth} 
    \centering
    \includegraphics[width=\linewidth]{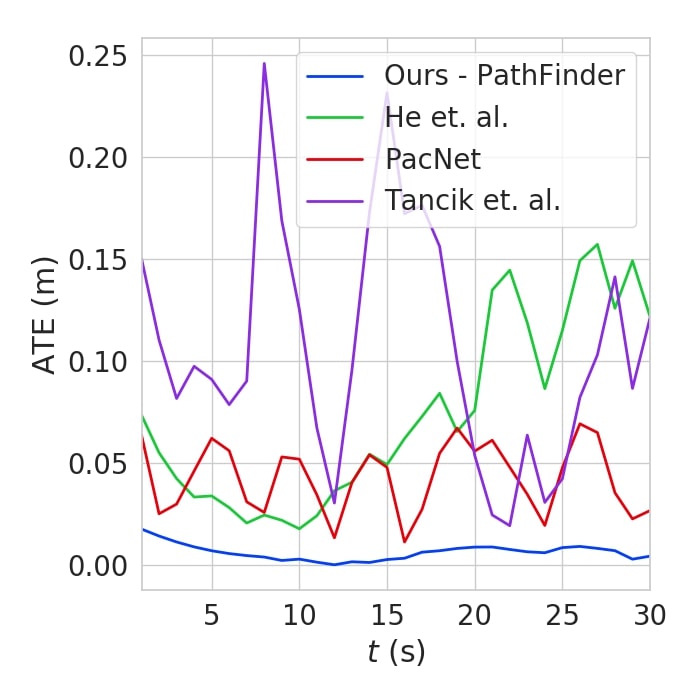}
    \label{fig:sub2}
  \end{subfigure}
  \begin{subfigure}{0.58\linewidth} 
    \centering
    \includegraphics[width=\linewidth, trim=1.2cm 0cm 1cm 0cm, clip]{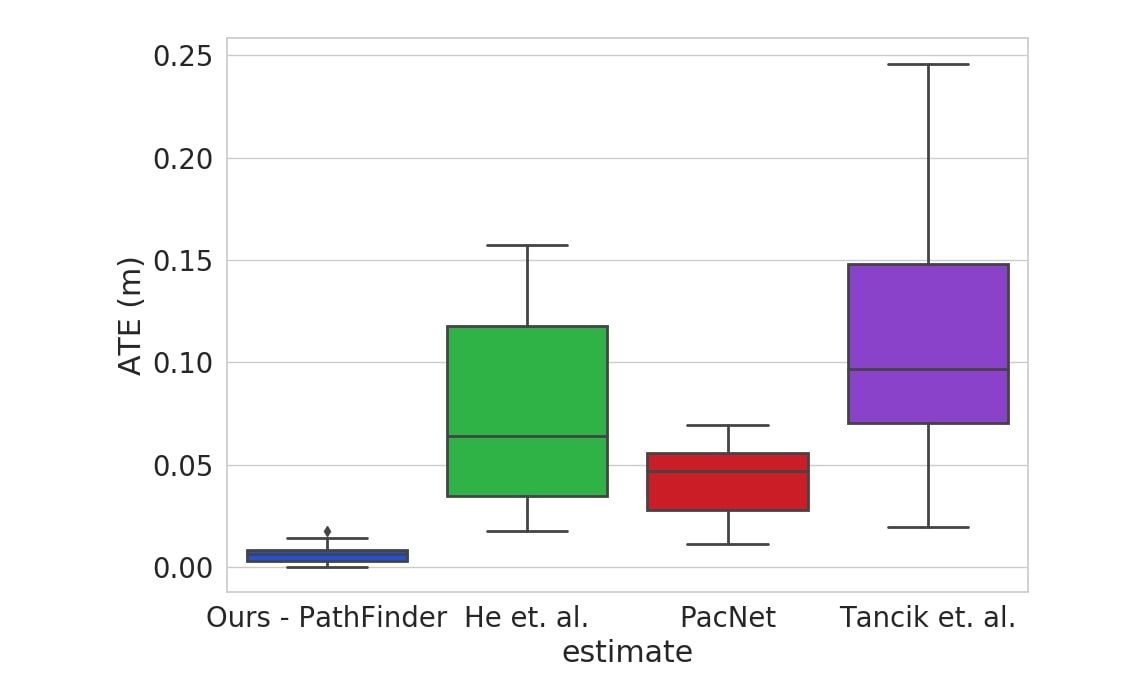}
    \label{fig:sub3}
  \end{subfigure}
 \vspace{-3mm}
  \caption{
\textit{Top plot:} Ground-truth trajectory and estimates from our method and baseline methods. \textit{Bottom plots:} The corresponding Absolute Trajectory Error (ATE) vs. time and ATE box plot for our method and baseline methods.}
  \label{fig:combined}
\end{figure}

In the top plot of Fig.~\ref{fig:combined}, we compare one estimated trajectory from our method to the ground-truth trajectory and trajectories estimated by the PAC-Net, He et al., and Tancik et al. methods. The estimate from our method closely aligns with the ground-truth trajectory, whereas the other estimates exhibit greater deviations from ground-truth. 
Moreover, our method produces consistently accurate position estimates over time, even when the camera movements are aggressive, whereas the other methods (which use a stationary camera) exhibit large variations in accuracy over time. 
This is evident from the bottom plots of Fig.~\ref{fig:combined}, which show the absolute trajectory error (ATE) over time and a box plot of the ATE for each method. Our method exhibits a consistently low ATE, while the ATEs of the other methods are higher and display large fluctuations.



\newlength{\subfigwidth}
\setlength{\subfigwidth}{0.24\textwidth} 

\begin{figure*}[t!]
    \centering
    \begin{subfigure}{\subfigwidth}
        \centering
        \includegraphics[width=\linewidth, trim=0.5cm 0.5cm 0.5cm 1.2cm, clip]{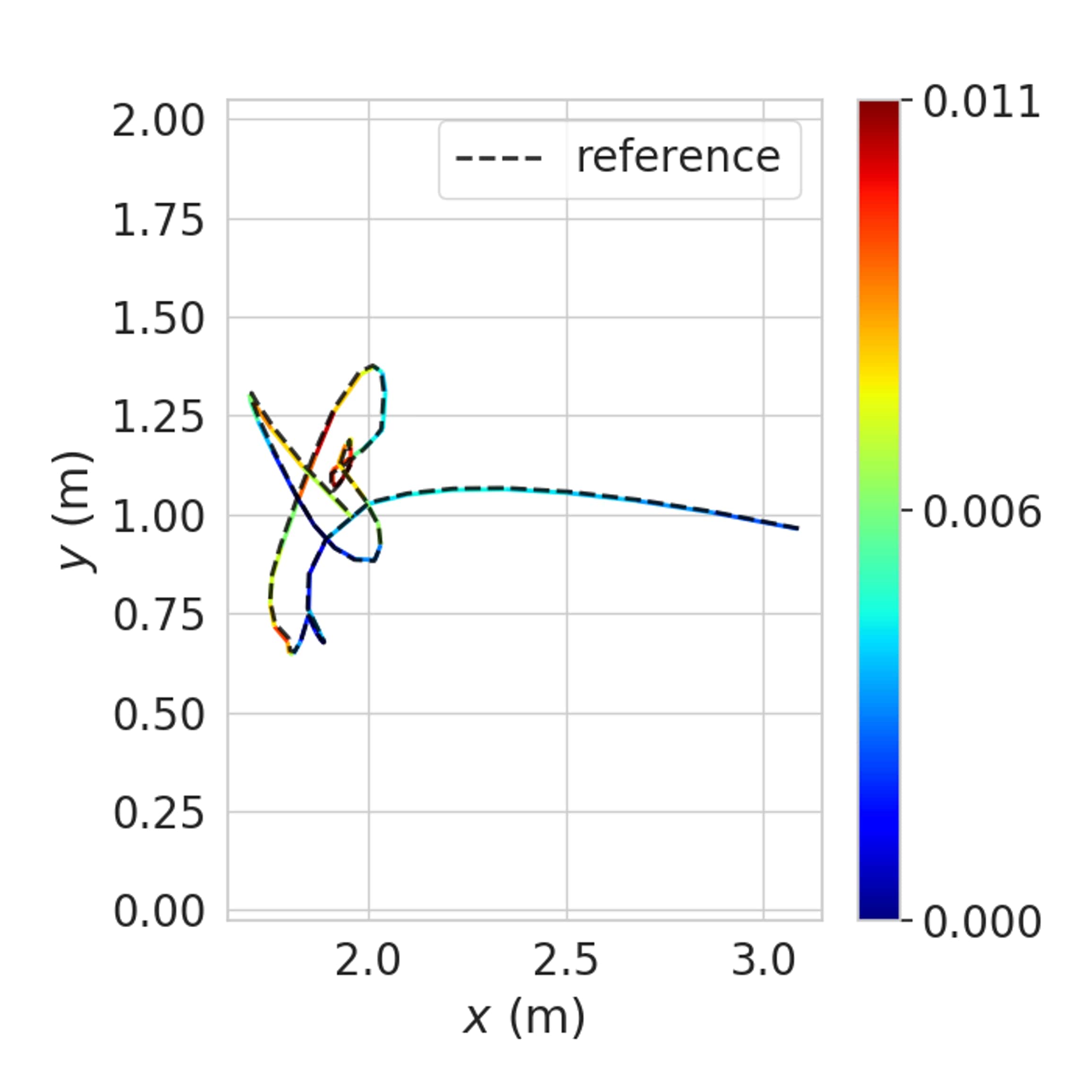}
        \label{fig:sub1}
    \end{subfigure}%
    \begin{subfigure}{\subfigwidth}
        \centering
        \includegraphics[width=\linewidth, trim=0.5cm 0.5cm 0.5cm 1.2cm, clip]{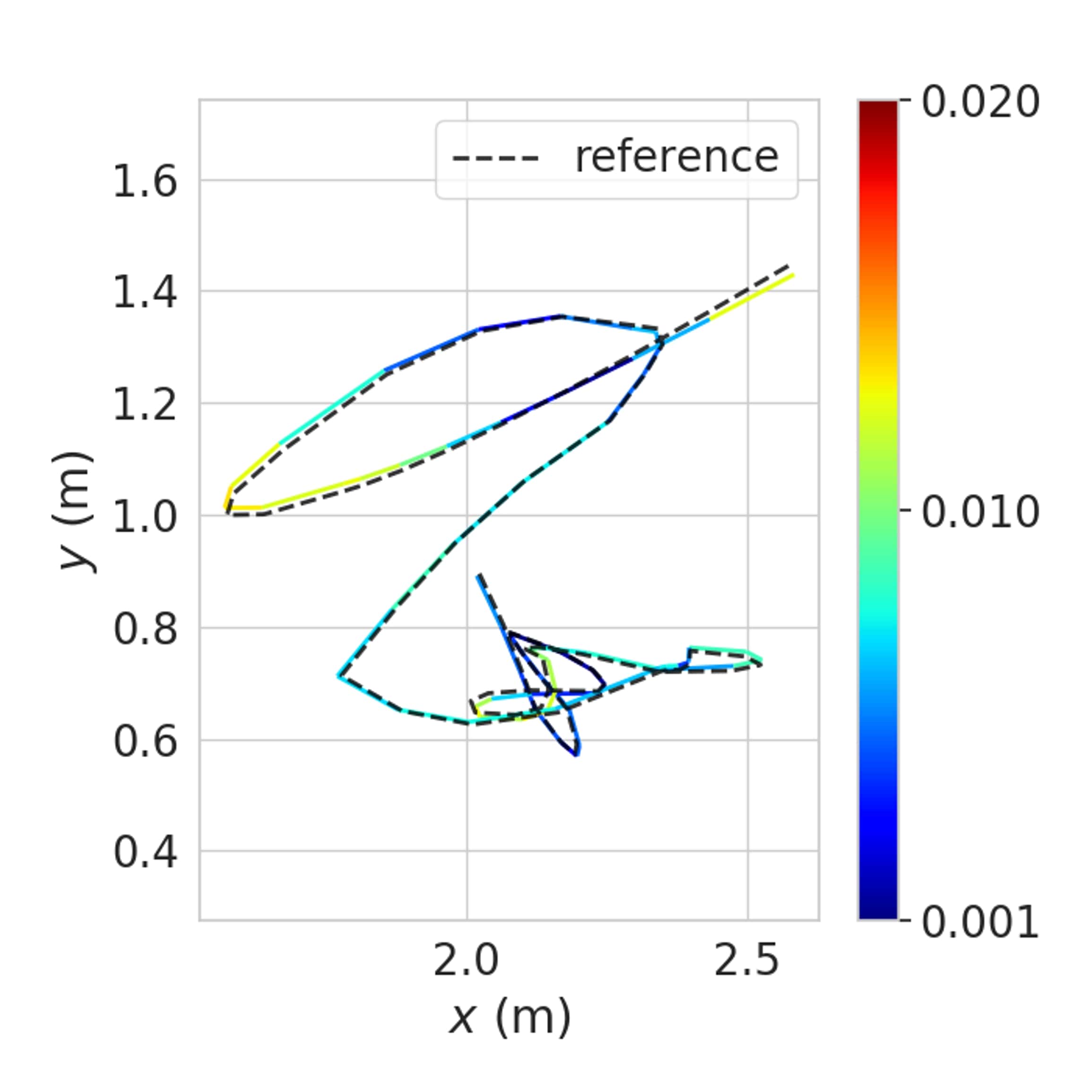}
        \label{fig:sub2}
    \end{subfigure}%
    \begin{subfigure}{\subfigwidth}
        \centering
        \includegraphics[width=\linewidth, trim=0.5cm 0.5cm 0.5cm 1.2cm, clip]{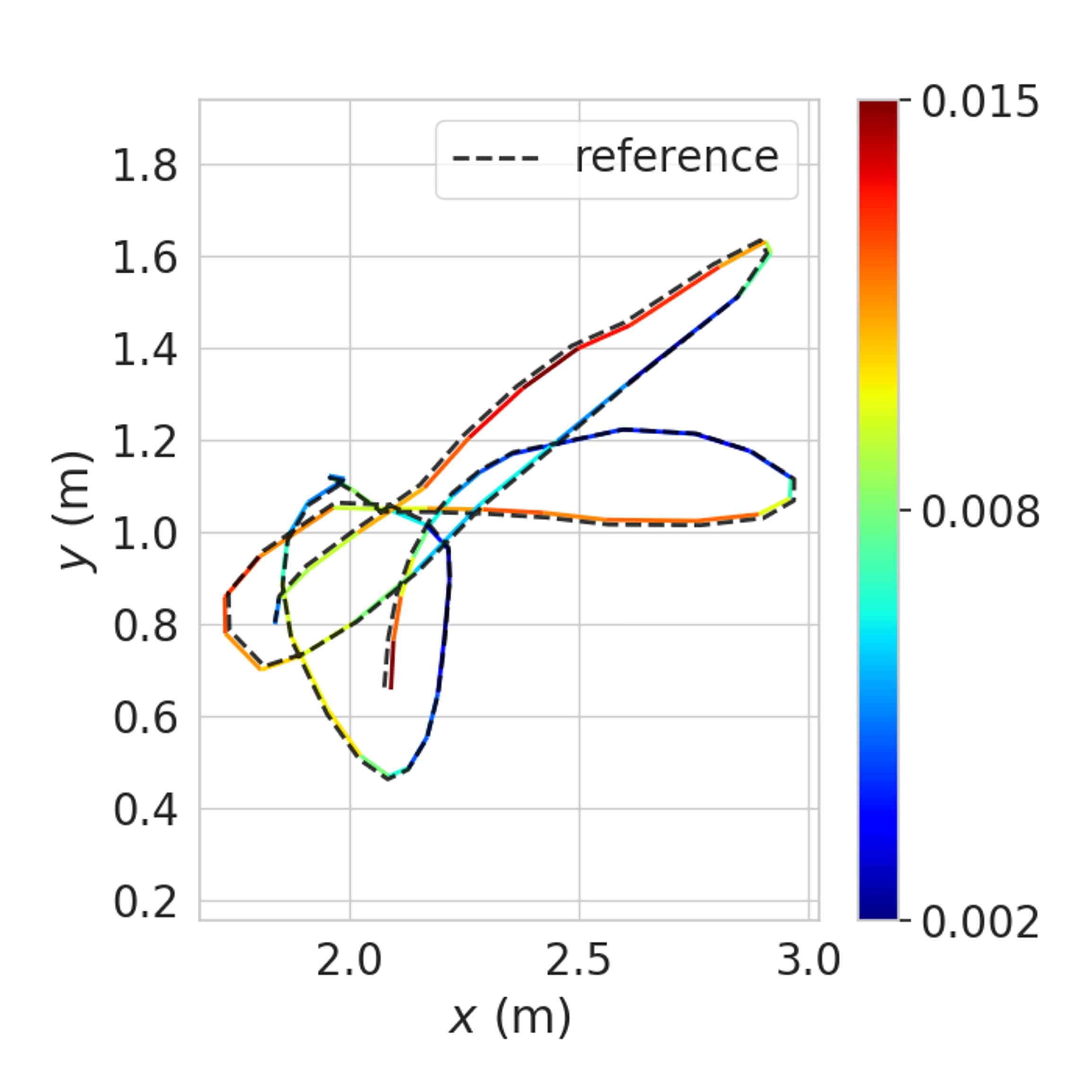}
        \label{fig:sub3}
    \end{subfigure}%
    \begin{subfigure}{\subfigwidth}
        \centering
        \includegraphics[width=\linewidth, trim=0.5cm 0.5cm 0.5cm 1.2cm, clip]{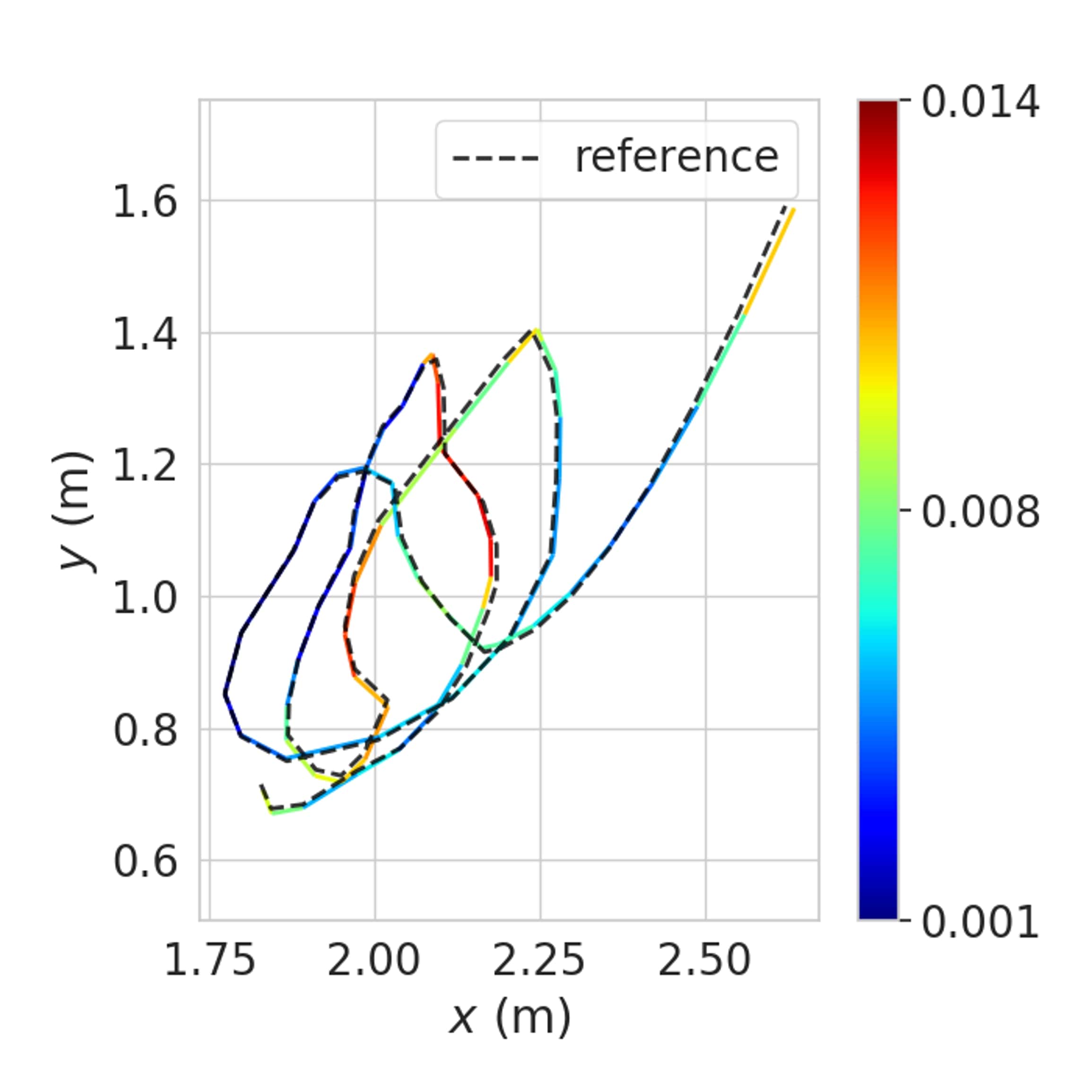}
        \label{fig:sub4}
    \end{subfigure}
    \vspace{-6mm}
    \caption{Ground-truth trajectories ({\it dashed lines}) and corresponding trajectories estimated by our method ({\it multicolored lines}), with the color 
    indicating the RMSE (m) between the ground-truth position $\mathbf{X}(t)$ and estimated position $\mathbf{X}'(t)$ at a time $t$.
    }
    \label{fig:trajectory}
    
\end{figure*}

\subsection{Effect of Number of Planes on Performance}
When the drone moves through the environment in our real-world trials, its camera captures numerous planes, but most of them tend to be small (trivial) and do not help improve tracking performance.
Intuitively, we know that packing more 
planes per sequence in our NLOS-Patch network scales up the cost of attention if we keep the hidden dimension of the transformer constant. Using our real-world data, we studied the effect on $\textsc{Rmse}_x$ of 
packing different numbers of masked planes into a sequence, 
without changing the hidden dimensions. 
The results of this ablation study 
are plotted in Fig. \ref{fig:planes_ablative}, which shows that using three planes results in the lowest $\textsc{Rmse}_x$ value.
Therefore, we chose to use the largest three planes as a suitable trade-off between performance and cost of attention.

\begin{figure}[t!]
\centering
\begin{subfigure}[b]{0.37\columnwidth}
  \centering
  \begin{adjustbox}{width=\linewidth}
    \input{sec/chart1}
  \end{adjustbox}
  \caption{Effect of number of masked planes in a single sequence on $\textsc{Rmse}_x$.}
  \label{fig:planes_ablative}
\end{subfigure}%
\hfill
\begin{subfigure}[b]{0.58\columnwidth}
  \centering
  \renewcommand{\arraystretch}{1.86} 
  \resizebox{\linewidth}{!}{%
  \begin{tabular}{@{} lcc @{}}
\toprule
& \text{$\textsc{Rmse}_x$} (mm) & \text{$\textsc{Rmse}_v$} (mm/s) \\
\midrule
Rotation (y) & 37.27 $\pm{5.35}$ & 1.89 $\pm{0.76}$ \\
Translation (x) & 22.12 $\pm{2.46}$ & 1.48 $\pm{0.51}$ \\
Translation (y) & 19.74 $\pm{2.38}$ & 1.04 $\pm{0.42}$ \\
\bottomrule
\end{tabular}}%
  \caption{Performance of our method with camera rotation about the y axis and translation along the x and y axes.}
  \label{results2}
\end{subfigure}
\caption{Ablation study results.}
\label{fig:comparison}
\end{figure}

\subsection{Effect of Camera Motion on Performance}


We assessed our method's 
performance under abrupt camera movements through tests in which a camera assembly, consisting of an Intel RealSense camera fixed between a Jetson Nano board and a LiPo battery, was manually moved in our real-world NLOS imaging setup. For motion capture tracking, 18 IR markers were attached to the camera assembly. 
Given the coordinate system in Fig. \ref{fig:dataset1}(b), in one trial, the camera was rapidly rotated about the y-axis at a rate of one full revolution per second for 10 s,
and in the other two trials,
the camera was translated back and forth 1 meter along either the x or y axis at about 1 m/s for 10 s. 

Sudden camera rotation disrupts the feature detector, causing inaccurate difference images. However, our dual-stream network architecture mitigates the effect of noisy or sparse difference image data on its position estimates. The table in Fig. \ref{results2} shows relatively low values of $\textsc{Rmse}_x$ and $\textsc{Rmse}_v$ for each test, 
indicating satisfactory performance. 


\subsection{Effect of Camera Sensor on Performance} 

To investigate the impact of camera sensor characteristics on our method's performance, we collected and analyzed raw color images from four different cameras: Sony A6000 Mirrorless, Intel RealSense D435i, IDS UI-3250CP-M-GL, and IMX214. These cameras vary in specifications such as megapixels, image resolution, FPS, and shutter types (rolling and global). The Intel RealSense and IMX214 cameras were mounted on drones, while the Sony A6000 and IDS cameras were operated manually. To enhance data integration, we paired raw images with IMU readings, obtained either directly from the cameras or through external IMUs. We calibrated the camera-IMU pairs using the Kalibr toolbox \cite{rehder2016extending}, following established methodologies to accurately determine the extrinsic relationships.

We tested our method in our real-world setup with a 1-minute image sequence captured using each of the four cameras. 
As shown in Table \ref{results1}, the tracking performance was similar across all four cameras. This consistency can be attributed to our NLOS-Patch network's ability to process planes of varying sizes and aspect ratios through example packing, as detailed in Section \ref{nlospatchnet}.

\begin{table}[t!]
\centering
\resizebox{\linewidth}{!}{%
\begin{tabular}{@{}cccccc@{}}
\toprule
 & \makecell{Sony\\A6000} & \makecell{Intel\\D435i} & \makecell{IDS\\camera} & \makecell{IMX214\\camera} \\
\midrule
$\textsc{Rmse}_x$ (mm) & \textbf{13.46 $\pm{2.24}$}  & 13.92 $\pm{2.73}$ & 14.63 $\pm{2.82}$ & 15.41 $\pm{2.56}$ \\
$\textsc{Rmse}_v$ (mm/s) & 1.84 $\pm{0.45}$ & 1.47 $\pm{0.39}$ & \textbf{1.21 $\pm{0.35}$} & 1.38 $\pm{0.44}$ \\
\bottomrule
\end{tabular}%
}
\caption{Tracking performance of our method, tested on different rolling shutter and global shutter cameras. 
}
\label{results1}
\vspace{-5mm}
\end{table}

%% file: sec/chart1.tex
\begin{tikzpicture}
    \begin{axis}[
            title={},
            title style={font=\bfseries\Large},
            xlabel={Number of Planes},
            ylabel={RMSE (\si{\milli\meter})},
            label style={font=\Large},
            xmin=0.5, xmax=10.5,
            ymin=0, ymax=6.5,
            xtick={2,4,6,8,10},
            ytick={1,2,3,4,5,6},
            xticklabels={2,4,6,8,10},
            yticklabels={1.0,2.0,3.0,4.0,5.0,6.0},
            tick label style={font=\large},
            legend pos=north west,
            legend style={font=\large, draw=none, fill=none},
            legend cell align={left},
            ymajorgrids=true,
            grid style={line width=.2pt, dashed, gray!40},
            every axis plot/.append style={line width=1.5pt},
            mark options={solid, scale=1.8},
            axis line style={-stealth, thick, black},
            tick style={black},
            legend columns=-1,
            legend style={/tikz/column 2/.style={column sep=10pt}},
        ]

       \addplot[color=blue!80!black, mark=square*, mark options={scale=1.5, fill=red}] coordinates {
               (1,1.9)
               (2,1.77)
               (3,1.59)
               (4,1.72)
               (6,3.9)
               (8,4.6)
               (10,5.5)
            };
        \addlegendentry{RMSE}
        \addlegendentry{RMSE}

    \end{axis}
\end{tikzpicture}

%% file: sec/6_conclusion.tex
\section{Conclusion}
\label{sec:conclusion}
We have proposed a novel data-driven approach for dynamic non-line-of-sight (NLOS) tracking using a mobile robot equipped with a standard RGB camera. To the best of our knowledge, this is the first NLOS tracking approach to handle dynamic camera environments. Our method leverages attention-based neural networks to accurately estimate the 2D trajectory of an occluded person in real-world Manhattan environments. Our NLOS tracking pipeline includes 
a plane extraction procedure 
that analyzes images from a moving camera to identify planes carrying maximum NLOS information. We 
employ novel transformer-based networks to process successive frames and estimate 
the person's position. 
The networks are capable of simultaneously processing multiple flat relay walls of different aspect ratios, which enhances the 
tracking performance.
We validated our approach on both synthetic and real-world datasets, demonstrating state-of-the-art results in dynamic NLOS tracking. Our method achieved an average positional 
RMSE of 15.94 mm on real data, outperforming existing passive NLOS methods and highlighting its effectiveness in practical scenarios.  
We plan 
to investigate the use of additional sensor data to further improve the accuracy and robustness of NLOS tracking.

